\documentclass[journal]{IEEEtran}
\usepackage{blindtext}
\usepackage{graphicx}
\usepackage[american]{babel}
\usepackage[justification=centering]{caption}
\usepackage{graphicx}
\usepackage{booktabs}
\usepackage{adjustbox,lipsum}
\usepackage{url}
\usepackage{algorithm}
\usepackage{multirow}
\usepackage[noend]{algpseudocode}
\usepackage{times}
\usepackage{amsmath}
\usepackage{amssymb}
\usepackage{tikz}
\usepackage{subfigure}
\usepackage{verbatim}
\usepackage{csquotes}
\usepackage{balance}
\usetikzlibrary{shapes,arrows}
\ifCLASSINFOpdf
\else
\fi
\begin{document}
\title{Semi-Supervised Self-Growing Generative Adversarial Networks for Image Recognition}

\author{
Haoqian Wang, Member, IEEE,
Zhiwei Xu,
Jun Xu, Member, IEEE\\
Wangpeng An,
Lei Zhang, Fellow, IEEE，
and Qionghai Dai, Senior Member, IEEE
\thanks{This project is partially supported by the National Natural Scientific Foundation of China (NSFC)
under Grant No. 61571259 and 61531014, in part by the Shenzhen Science and Technology Project under Grant (GGFW2017040714161462, JCYJ20170307153051701)}
\thanks{H. Q. Wang, Zhiwei Xu and W. P. An are with the Graduate School at Shenzhen, Tsinghua University, and slso with  Shenzhen Institute of Future Media Technology, Shenzhen 518055, China (e-mail: wanghaoqian@tsinghua.edu.cn, anwangpeng@gmail.com).}
\thanks{J. Xu is with Media Computing Lab, College of Computer Science, Nankai University, Tianjin, China. (e-mail: nankaimathxujun@gmail.com).}
\thanks{Q. Dai is with TNLIST and Department of Automation, Tsinghua University, Beijing 100084, China (e-mail: qhdai@tsinghua.edu.cn). }
\thanks{L. Zhang is with the Department of Computing, The Hong Kong Polytechnic
University, Hong Kong (e-mail: cslzhang@comp.polyu.edu.hk).}}
\maketitle
\begin{abstract}
Image recognition is an important topic in computer vision and image processing, and has been mainly addressed by supervised deep learning methods, which need a large set of labeled images to achieve promising performance. However, in most cases, labeled data are expensive or even impossible to obtain, while unlabeled data are readily available from numerous free on-line resources and have been exploited to improve the performance of deep neural networks. To better exploit the power of unlabeled data for image recognition, in this paper, we propose a semi-supervised and generative approach, namely the semi-supervised self-growing generative adversarial network (SGGAN). Label inference is a key step for the success of semi-supervised learning approaches. There are two main problems in label inference: how to measure the confidence of the unlabeled data and how to generalize the classifier. We address these two problems via the generative framework and a novel convolution-block-transformation technique, respectively. To stabilize and speed up the training process of SGGAN, we employ the metric Maximum Mean Discrepancy as the feature matching objective function and achieve larger gain than the standard semi-supervised GANs (SSGANs), narrowing the gap to the supervised methods. Experiments on several benchmark datasets show the effectiveness of the proposed SGGAN on image recognition and facial attribute recognition tasks. By using the training data with only $4\%$ labeled facial attributes, the SGGAN approach can achieve comparable accuracy with leading supervised deep learning methods with all labeled facial attributes.
\end{abstract}

\IEEEpeerreviewmaketitle

\section{Introduction}
\label{sec:1}
In the past decade, we have witnessed the increasing interests in the image recognition problem solved by the deep learning approaches~\cite{krizhevsky2012imagenet,Simonyan14c,he2016deep}. This interest is expanding quickly to many different fields ever since the advent of deep convolution neural networks~\cite{krizhevsky2012imagenet,Simonyan14c,goodfellow2014generative,he2016deep}, resulting in many effective approaches
in many different computer vision fields~\cite{tip_video,tip_track,tip_reid,tip_fcn, tip_saliency}. However, despite these exciting progresses, most existing approaches are supervised learning based and largely limited by resorting to huge amounts of data with labels. Labeling these data will incur expensive costs on human labor. To alleviate the dependence of supervised learning approaches on the labeled data, many semi-supervised learning approaches~\cite{semi-rgbd,tip_small_data,tip_semi_face_recognition,cherniavsky2010semi,XU2019679} have been developed to exploit the power of the numerous unlabeled data available in free on-line resources for the image recognition problem. On the other hand, with the successes of Deep Convolutional Generative Adversarial Networks (DCGAN)~\cite{DBLP:journals/corr/RadfordMC15} on general pattern recognition tasks, Generative Adversarial Networks (GANs) have been widely applied into unsupervised learning problems~\cite{improvegan}.

It is well known that the GANs can hardly be trained deeply enough when compared to the other concurrently networks such as ResNet~\cite{he2016deep}. This is because that the generator of the GANs are usually very shallow and can often drift to \enquote{model collapse} (a parameter setting where it always emits the same point), restricting the GANs to grow up to achieve promising performance on large scale datasets such as ImageNet~\cite{imagenet_cvpr09}. In this paper, we propose a novel a self-growing GAN (SGGAN) for large scale image recognition tasks. The proposed SGGAN is a united semi-supervised GAN containing three self-growing groups. Each group contains a generator and a discriminator, which are trained at the same time and compete against each other to reach the Nash equilibrium of the game theory through an adversarial objective~\cite{goodfellow2014generative}. The generator is trained to defeat the discriminator by creating virtually realistic images (maximize the loss), and the discriminator is trained to distinguish the images generated by the generator  (minimize the loss). Through this min-max game, the loss of generator will become increased while the loss of the discriminator will becomes decreased. Finally, the two losses will become closer to each other, and reach an equilibrium in the end. 

In semi-supervised learning (SSL) framework, label inferring is a major challenging to its success. Given an amount of labeled data and a larger amount of unlabeled data, the SSL framework can infer the latent label information of the unlabeled data from the labeled data by considering the structures and distributions of all these data as a whole. In order to guarantee the success of the semi-supervised learning approach, label inference of the unlabeled data is the most significant problem to address. For the labels assigned to the unlabeled data, the false positive rate of the label inference process is more important than the true negative rate for the recognition performance, since false positive labels would add noise into the training data and thus make the training unstable. Therefore, the confidence of the latent labeled data (i.e., unlabeled data with latent labels) should be accurate enough. Moreover, the semi-supervised classifiers may not improve if they perform well on the same types of data. Therefore, the two main obstacles in label inference are: how to measure the confidence of the unlabeled data, and how to generalize the semi-supervised learning classifier. In this paper, we propose to address the first problem through threshold setting techniques~\cite{Chapelle:2010:SL:1841234}, in which only the unlabeled data with recognition probability larger than a pre-set threshold will be assigned with a latent label. We solve the second problem by proposing a novel technique named convolution-block-transformation (CBT) proposed by us. In our proposed network, the depth is designed to be deep in order to generalize the classifier since deeper model enables the network to learn more information from the unlabeled data than the shallower one. It is difficult to directly train a deep network in our case, so we propose a simple yet effective convolution block transformation (CBT) technique to transfer weights from a shallower network to a deep one by shortcut and an adaptive scaling layer following the shallower convolution block. We evaluate our method on CIFAR10, SVHN and face attribute recognition dataset, which is more challenging due to complex face variations.

In summary, the major contributions of this paper are summarized as follows:
\begin{itemize}
\item We propose an semi-supervised self-growing generative adversarial network (SGGAN) for image recognition problem. We handle the semi-supervised learning problem via label inference to improve the performance of the training network. 
\item We introduce the minimum mean discrepancy (MMD) as the objective of the feature matching stage to replace the traditional $\ell_1$ distance objective. The employed MMD can help to stabilize the  training of the proposed SGGAN model, and thus avoid the model collapse pitfall of traditional GANs.
\item We propose a novel convolution block transformation (CBT) technique to harmonize the self-growing process of the proposed SGGAN model to address the generalization of its classifier. We prove it is easier to train a model growing from a shallow network to a deep one, and thus achieving better performance.
\item We conduct extensive experiments on image and face attribute recognition problems to systematically evaluate our proposed SGGAN model. We demonstrate that MMD and CBT can separately and simultaneously stabilize the training of the proposed SGGAN. When compared with supervised methods, SGGAN can achieves competitive or even higher accuracies on various benchmark datasets when compared with state-of-the-art GAN based approaches such as the Improved GAN~\cite{improvegan} and supervised learning networks such as VGG-16~\cite{Simonyan14c} and ResNet-50~\cite{he2016deep}.
\end{itemize}
The rest of this paper is organized as follows: In Section~\ref{sec:related_work}, we briefly reviews related work on semi-supervised learning, generative adversarial networks, the optimization of GAN and face attribute recognition. In Section~\ref{sec:method}, we introduces the architecture of our proposed semi-supervised self-growing generative adversarial network and how to train our SGGAN. Experiments and detailed analysis are introduced in Section~\ref{sec:results}. Finally, we conclude this paper in Section~\ref{sec:conclusion}.

\begin{figure*}[htb]
    \centering
    \includegraphics[height =0.4\textheight, width=0.8\textwidth,angle=0]{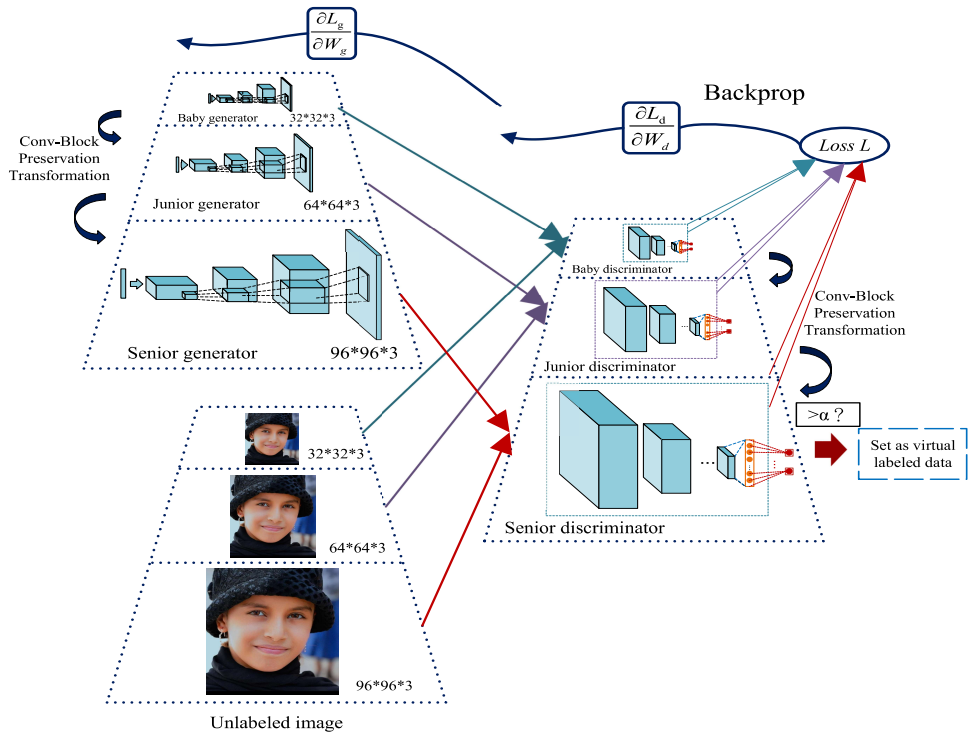}
    \caption{The architecture of our semi-supervised self-growing generative adversarial network (SGGAN). SGGAN starts from the basic baby generator and discriminator, in which the junior and senior generator/discriminator are self-grown from the baby counterparts via our proposed CBT technique.}
\label{fig:framework}
\end{figure*}
\section{Related Work}
\label{sec:related_work}
\subsection{Semi-Supervised Learning} Semi-supervised algorithm~\cite{Chapelle:2010:SL:1841234} falls between unsupervised learning (e.g., clustering) and supervised learning (e.g., classification or regression) on providing the data labels~\cite{newDNAProt,cvdp,mcwnnm,twsc,pgpd,gid2018,xu2018thesis,xu2018real,
xu2019nac,xuaccv2016,An_2018_CVPR,Liang_2018_CVPR,rssc,nrc,tstss,xu2019star,hou2019nlh,RANet2019}. Semi-supervised learning~\cite{zhu2009introduction} contains multiple types of training strategy, such as self-training~\cite{rosenberg2005semi} and co-training~\cite{zhou2005semi}. Recently, Zhuang et al.~\cite{tip_semi} considered the label information in the graph learning stage. Specifically, they enforce to be zero the weight of edges between every two labeled samples from different classes. To make use of the unlabeled data, one simple and effective way is to predict the labels of the unlabeled data by employing the model trained on existing labeled data. Indeed, the premise behind semi-supervised learning is that the learned statistics in the labeled examples contain information which is useful to predict the unknown labels. Self-training~\cite{rosenberg2005semi} is one of the earliest semi-supervised learning strategy using unlabeled data to improve the training of recognition systems. The high confidence that the model predicts against a sample indicates the high probability of correct prediction.
\begin{table*}[htb]
\begin{minipage}{0.5\linewidth}
\scriptsize
\centering
\caption{Architecture of Discriminators.}
\begin{tabular}{ccc} 
 \toprule
 \textbf{Baby D} & \textbf{Junior D} & \textbf{Senior D} \\ 
 \midrule
 Input (32$\times$32$\times$3) & Input (128$\times$128$\times$3)
 & Input (512$\times$512$\times$3) \\ 
 \midrule
 Conv3-64S1 & Conv3-64S1 & Conv3-64S1 \\ 
 Conv3-64S1 & Conv3-64S1 & Conv3-64S1 \\ 
 Conv3-64S2 & Conv3-64S2 & Conv3-64S2 \\ 
 \midrule
 Conv3-128S2\textbf{$\times2$} & Conv3-128S2\textbf{$\times2$} & Conv3-128S2\textbf{$\times2$} \\ 
 Conv3-128S1\textbf{$\times2$} & Conv3-128S1\textbf{$\times2$} & Conv3-128S1\textbf{$\times2$} \\ 
 Conv3-128S1\textbf{$\times1$}  & Conv3-128S1\textbf{$\times1$} & Conv3-128S1\textbf{$\times1$} \\ 
 \midrule
 $- $ & Conv3-192S1\textbf{$\times2$} & Conv3-192S1\textbf{$\times2$} \\ 
 $- $ & Conv3-192S1\textbf{$\times2$} & Conv3-192S1\textbf{$\times2$} \\ 
 $- $ & Conv3-192S2\textbf{$\times1$} & Conv3-192S2\textbf{$\times1$} \\ 
 \midrule 
 $- $ & $- $ & Conv3-256S1\textbf{$\times2$} \\ 
 $- $ & $- $ & Conv3-256S1\textbf{$\times2$} \\ 
 $- $ & $- $ & Conv3-256S2\textbf{$\times2$} \\ 
 \midrule 
 \multicolumn{3}{c}{Dropout(0.5)} \\
 \midrule 
 \multicolumn{3}{c}{Global Average Pooling} \\
 \midrule 
 \multicolumn{3}{c}{FC} \\
 \midrule 
 \multicolumn{3}{c}{softmax} \\
 \bottomrule
\end{tabular}
\label{tab:discriminator}
\end{minipage}
\begin{minipage}{0.5\linewidth}
\scriptsize
\centering

\caption{Architecture of Generators.}
\begin{tabular}{ ccc } 
 \toprule
 \textbf{Baby G} & \textbf{Junior G} & \textbf{Senior G} \\ 
 \midrule 
 \multicolumn{3}{c}{Sample 100 number from Uniform Distribution} \\
 \midrule 
 \multicolumn{3}{c}{FC-512*4*4} \\
 \midrule 
 \multicolumn{3}{c}{Reshape-(4,4,512)} \\
 \hline
Deconv5-256S2 & Deconv5-256S2 & Deconv5-256S2 \\ 
Deconv5-128S2 & Deconv5-128S2 & Deconv5-128S2 \\ 
 Deconv5-128S2 & Deconv5-128S2 & Deconv5-128S2 \\ 
 - & Deconv5-128S1 & Deconv5-128S1 \\ 
 \midrule 
- & 
Deconv5-128S2 & 
Deconv5-128S2 \\ 

- & Deconv5-128S2 & Deconv5-128S2 \\ 

- & 
Deconv5-128S1 & Deconv5-128S1 \\  
\midrule 
$-$ & Deconv5-64S2 & 
Deconv5-64S2 \\ 

$-$ & Deconv5-64S2 & Deconv5-64S2 \\ 

$-$ & Deconv5-32S1 & Deconv5-32S1 \\ 
\midrule 
$- $ & $-$ & Deconv5-32S2 \\ 
\midrule 
$- $ & $- $& Deconv5-32S2 \\ 
\midrule 
Output(32$\times$32$\times$3)& Output(128$\times$128$\times$3) & Output(512$\times$512$\times$3) \\ 
\midrule 
\multicolumn{3}{c}{Tanh Activation} \\
\midrule 
\end{tabular}
\label{tab:generator}
\end{minipage}
\end{table*}
\subsection{Generative Adversarial Networks} The training objective of GANs is to find a Nash equilibrium between the discriminative and generator networks by a min-max game. Denote by the generative network in GAN by $G$ and the discriminative network in GAN by $D$. The purpose of the $G$ network is to generate virtually realistic images and the purpose of the $D$ network is to distinguish between the virtually generated and realistic unlabeled images through the min-max optimization problem. As described in the original paper~\cite{goodfellow2014generative}, the purpose of the generative modeling is to find a probabilistic model $Q$ that matches the true data distribution $P$. The training of GAN can be interpreted as minimizing the Jensen-Shannon divergence under some ideal conditions. The Jensen-Shannon divergence is not measured by the K-L divergence between $P$ and $Q$, i.e., $KL[P\|Q]$ or $KL[Q\|P]$, but is between the two extreme cases $KL[P\|Q]$ and $KL[Q\|P]$. And this property of the Jensen-Shannon divergence can push the generator to generate better samples than other methods~\cite{goodfellow2014generative}. Actually, Nash equilibrium is difficult to achieve and the assumptions behind GANs maybe too strong to perfectly match the cases in real-world applications. In the work of DCGAN~\cite{DBLP:journals/corr/RadfordMC15}, there are several techniques proposed to stabilize the training of the GANs, i.e., using leaky-ReLUs and batch normalization for the training of the discriminator network, and convolution with stride $2$ instead of max-pooling layers for the training of the generator network. These techniques work very well and have become a standard setup in recent GAN based approaches. Recently, Wasserstein distance~\cite{arjovsky2017towards} is introduced with theoretically proved effectiveness as the objective of the generative model to stabilize the training process of GAN. The main advantage of Wasserstein distance based GAN frameworks is that this distance can guarantee great stability for training the generative model, which is not limited to the DCGAN approach.

Existing GAN based approaches can be categorized into two types from the perspective of their motivations. The first type is the divergence minimization based approaches~\cite{DBLP:journals/corr/RadfordMC15}, which mainly focus on designing an effective generator network to produce virtually realistic images, and treat the discriminator network as an auxiliary model. And the second type is the contrast function based approaches~\cite{improvegan}, which attempt to enhance the discriminating power of the discriminator by simulating a large amount of fake samples. Our work can be categorized into the second type of approaches.

\subsection{GAN based semi-supervised learning}
%There have been numerous studies that examine the use of generative models in a semi-supervised setting. 
Donahue et al.~\cite{adversarial_inference} introduced an adversarial formulation
with a third component, which they call the \enquote{encoder}. While the generator maps a simple latent distribution to data
space, the encoder attempts to encode real data to some latent space. They show that this encoder is capable of learning
to invert the generator, and can be used as a feature for a supervised training.
On the autoregressive side, Dai et al. ~\cite{semi_sequence} explored the
idea of first “pretraining” a sequence model to perform a
task on unlabeled text data. These pretrained weights are
then used to train supervised models for text classification.
Their results show improved learning stability and model generalization.
Radford et al.~\cite{generate_reviews} trained an mLSTM RNN on Amazon
reviews to learn a language model and then used its internal cell state from the last time step as features for the
subsequent supervised task of sentiment analysis of Amazon
reviews. This enabled the authors to match the state-of-the-art in their sentiment analysis dataset with significantly
less labeled data and to surpass it with the fully-supervised learning.
Recently, Salimans, et al.~\cite{improvegan} proposed a way to utilize GANs for a
classification task with $k$ classes. Specifically, they propose
an extension to the vanilla GAN where the labeled
dataset is augmented with samples from the generator. The
discriminator is also modified to predict $(k + 1)$th classes: the
original $k$ classes and a new class for fake (generated) data.
In a sense this helps the discriminative model by augmenting
a smaller labeled dataset with larger unlabeled set of real
examples and generated samples.
%The existing methods for semi-supervised learning using GANs modify the regular GAN discriminator to have $k$ outputs corresponding to $k$ real classes~\cite{catgan}, and in some cases a $(k + 1)$th output that corresponds to fake samples from the generator~\cite{adversarial_inference,improvegan,semi_gan}. The generator is mainly used as a source of additional data (fake samples) which the discriminator tries to classify under the $(k + 1)$th label.

\subsection{Face Attribute Recognition}
Face attribute recognition in the wild is a challenging problem due to complex face variations such as varying lightings, scales, and occlusions, etc. Traditionally, previous attribute recognition approaches~\cite{berg2013poof,bourdev2011describing,kumar2011describable} focus on extracting effective hand-crafted low-level features, e.g., edges, HSV, and gradients, etc, from the detected faces. Then the extracted features are fed into a standard classifier, such as SVM~\cite{svm} and random forest~\cite{random_forest}. For instance, the authors of FaceTracer~\cite{kumar2008facetracer} split the whole face region into multiple sub-regions, extracted multiple types of features for each region, and train a SVM classifier on the concatenated features. Recently, deep learning (especially CNN based) methods~\cite{krizhevsky2012imagenet} have achieved great success in face attribute recognition due to their ability to learn discriminative features from huge amount of labeled data. The authors in~\cite{celeba} applied two CNNs (ANet and LNet) to the face attribute recognition task, on which the LNet is trained to locate the entire face region and the ANet is trained to extract high-level face representation. Finally, the extracted features are fed into a SVM classifier to produce the final recognition results. In \cite{DBLP:journals/corr/RuddGB16}, the authors proposed a mixed objective to optimize $40$ face attributes together in a single CNN with $138$ million network parameters. However, these supervised deep learning methods are limited by largely depending on huge amount of labeled training data, which is very costly in real-world applications. This motivates us to utilize the large amount of freely available unlabeled data for the face attribute recognition in a semi-supervised manner.

\section{Semi-Supervised Self-Growing Generative Adversarial Network}
\label{sec:method}
In this section, we first reveal the mechanism of the proposed semi-supervised self-growing generative adversarial network (SGGAN) by presenting its architecture in details. Then the convolution block transformation strategy for network self-growing is illustrated. Finally, we introduce the MMD as an effective metric to stabilize the training of our model.

\subsection{Architecture of SGGAN}
The flowchart of the proposed SGGAN is illustrated in Figure~\ref{fig:framework}. Our SGGAN network includes a group of GANs, in which the junior generator or discriminator grows from corresponding baby counterpart, and the senior generator or discriminator grows from corresponding junior counterpart. The detailed description of the structures of three generators and three discriminators are listed in Table~\ref{tab:discriminator} and ~\ref{tab:generator}, respectively. The convolutional layer parameters are denoted as “(convolution type)(kernel size)-(number of channels)-S(stride)”. The activation functions we employed for the generator and discriminator are ReLU and Leaky-ReLU, respectively. Batch normalization is used after each convolution layer. The self growing process will be discussed in the next subsection. In the whole network, the fundamental component is named as the GAN cell, which is composed of a generator network and a discriminator network. %The GAN cell could function as an individual network regardless of the accuracy.

In the GAN cell, the discriminator is deeper and sometimes has more filters per layer than the corresponding generator. The reason is that it is important for the discriminator to be able to correctly estimate the ratio between the true data density and generated data density, but it may also be an artifact of the ``mode collapse'' since the generator tends not to use its full capacity with current training methods~\cite{gan_tutorial}. We introduce each component of the proposed SGGAN model as follows.
\subsubsection{Generator}
\begin{figure}[htbp]
\centering
\includegraphics[width=0.5\textwidth]{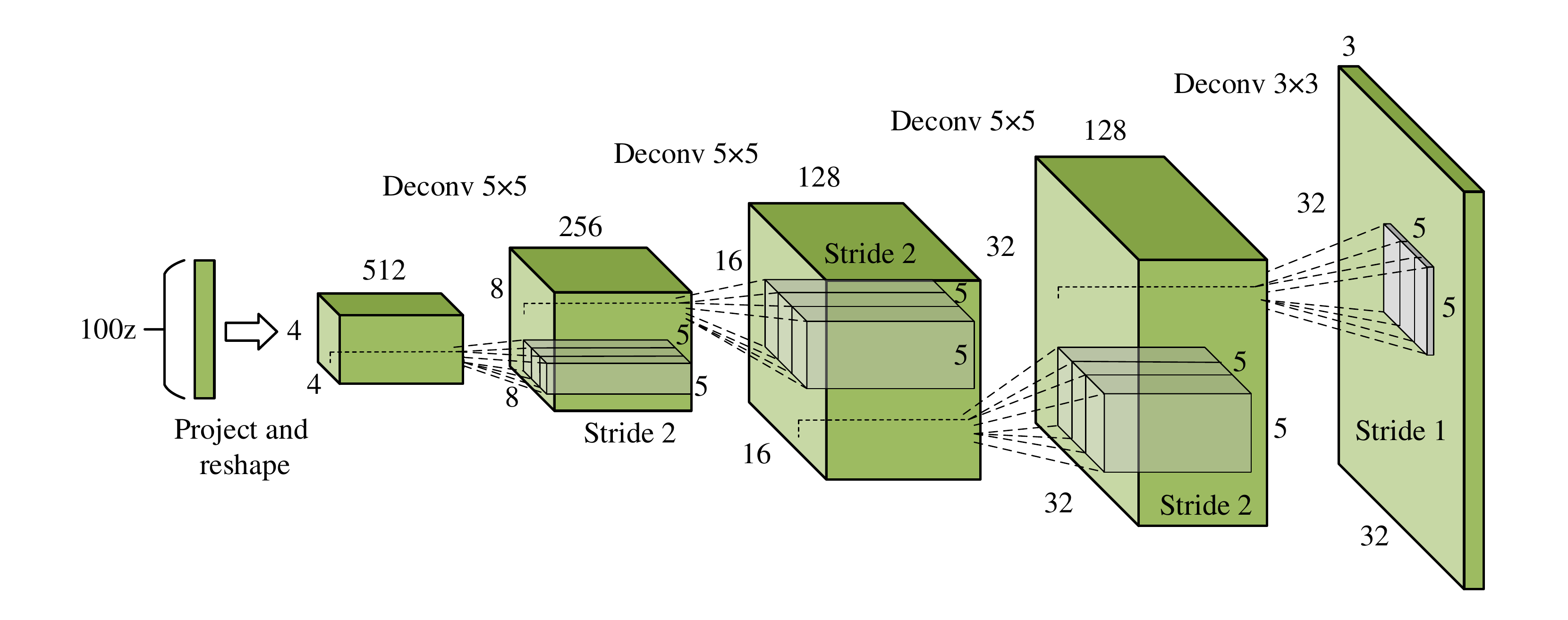}
\caption{The detailed architecture of Baby Generator.}
\label{fig:generator}
\end{figure}
The generator takes as input a random vector $z$ (drawn from a Gaussian distribution). After reshaping $z$ into a $4$-dimensional shape, it is fed to the generator that starts with a series of upsampling layers. Each upsampling layer represents a transposed convolution operation with a stride of $2$.  The transposed convolution work by swapping the forward and backward passes of a convolution. The upsampling layers go from deep and narrow layers to wider and shallower ones. The stride of a transposed convolution operation defines the upsampling factor of the output layer. With the stride of $2$, the size of output features will be twice that of the input layer. After each transposed convolution operation, the reshaped $z$ becomes wider and shallower. All transposed convolutions use a $5\times5$ kernel with depths reducing from $512$ to $3$, which indicating a RGB color image. The output of the final layer is a $H\times W\times3$ tensor, squashed between values of $-1$ and $1$ through the Hyperbolic Tangent ($tanh$) function. The shape of the final output is defined by the size of the training image. Specifically, if we train the generator on the SVHN dataset~\cite{svhn}, it will produce an image of size $32\times32\times3$.

\subsubsection{Discriminator}
\begin{figure}[htbp]
\centering
\includegraphics[width=0.5\textwidth]{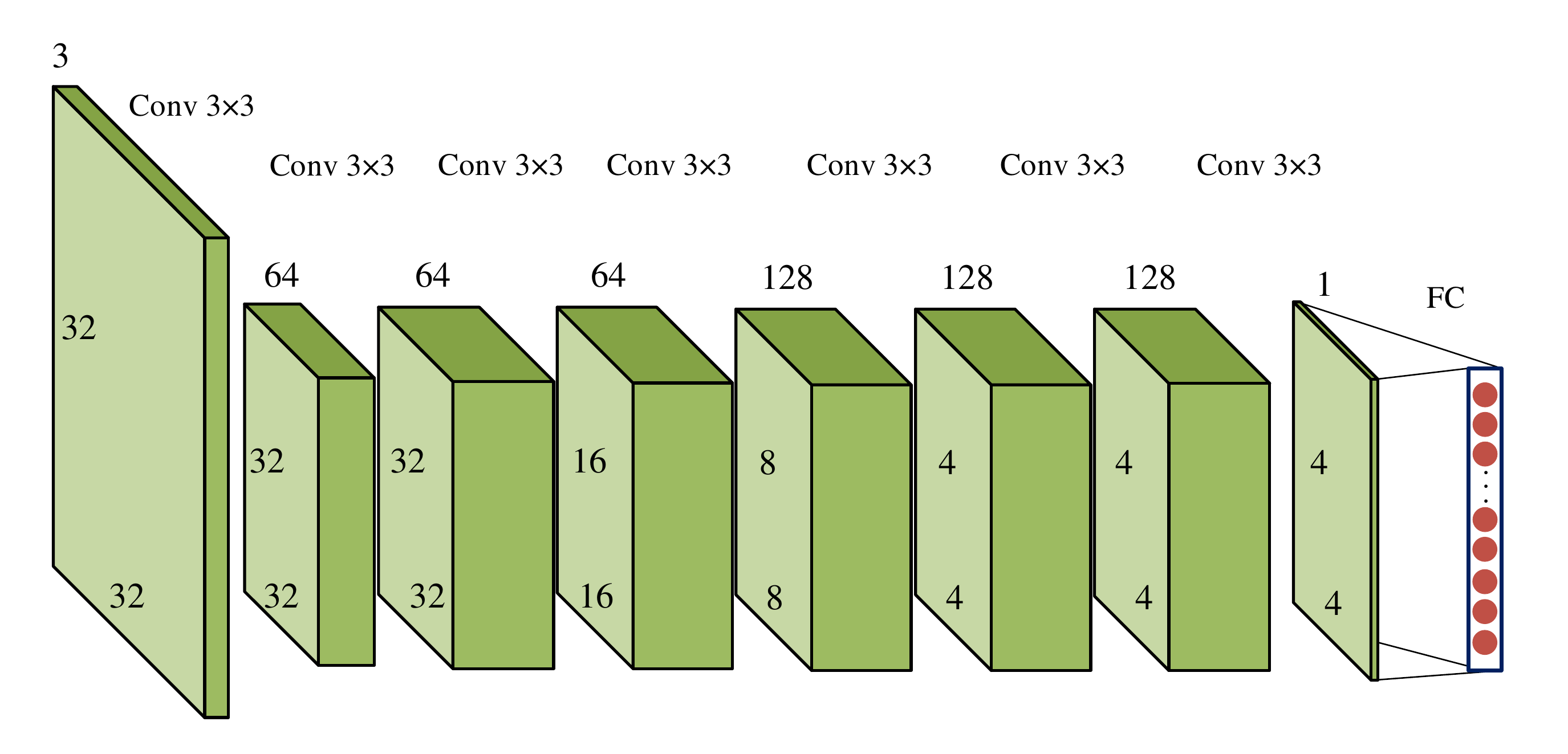}
\caption{The detailed architecture of Baby Discriminator.}
\label{fig:generator}
\end{figure}
The baby discriminator has $9$ CNN layers with Batch Normalization~\cite{batch_norm}, followed by Leaky-ReLU activation function. It is the same with the deep neural networks used for image recognition~\cite{tip_class}, object detection~\cite{tip_detect}, and image segmentation~\cite{tip_segm}, etc. The difference is that the Leaky-ReLU~\cite{leaky_relu} is used in our discriminator instead of the regular ReLU~\cite{relu}. The reason we employ Leaky-ReLU instead of the regular ReLU is that, the regular ReLU function will truncate the negative values to $0$, which will block the gradients to flow through the generative networks. Instead of forcing the negative part to be $0$, the Leaky-ReLU allows a small negative value to pass through the activation layer.  Theoretically, Leaky-ReLU represents an attempt to solve the dying ReLU problem~\cite{p_relu}. This situation occurs when the neurons do not move in a state in which ReLU units always output $0$s for all inputs. For these scenarios, the gradients do not flow back through the network. This problem is especially important for GAN since the only way the generator learns is by receiving the gradients from the discriminator. 
%Compared with ReLUs, the Leaky-ReLU is popular because it can help the gradients flow easier through the GAN architecture.The function computes the greatest value between the features and a small factor.

Our baby discriminator takes into a $32\times32\times3$ image tensor as input. Being opposite to the generator, the discriminator contains a series of convolutions with a stride of $2$. Each layer reduces the spatial dimensions of feature vector by reducing its size by half, along with doubling the number of learned filters. Given the training data from $k$ classes, the discriminator will output $(k+1)$ neurons to represent these $k$ classes, where the $(k+1)$th class demonstrates the generated images. We employ the softmax activation function as the output of the final layer to generate the confidence for samples from each class. When the discriminator captures the difference between the generated image and realistic image, it will send a signal to the generator counterpart. This signal is the gradient that flows backward from the discriminator to the generator. Once receiving this signal, the generator is able to adjust its parameters accordingly to generate latent data whose distribution is closer to the true data distribution than the previous generated ones. In the final stage, the generator will produce data as good as that the discriminator hardly distinguishes them apart.

\subsubsection{Label inference by discriminator}
The latent labels of our unlabeled images are firstly created via the baby discriminator, and then updated by the junior discriminator. The self-training approaches usually needs a threshold value to infer the latent labels. The threshold value of the confidence is determined on the validation dataset of CelebA dataset~\cite{celeba}; we set the threshold value as $0.98$ in our experiments. In this way, we can utilize more unlabeled images, and hence train deeper neural networks for better recognition performance.

\subsection{Self-Growing Network}
In this section, we propose a convolution block transformation (CBT) technique to transform an existing network into a deeper one. Our idea is motivated by the Net2Net model~\cite{DBLP:journals/corr/ChenGS15}. In Net2Net, Chen et al. proposed to initialize a bigger model using the weights of a smaller model. However, they only initialize the weights of one layer in each cycle, and this operation has difficulties with the batch normalization (BN) layer. This is because that the BN layer requires running forward inference on the training data to calculate the mean and variance of activation function, which are then used to set the output scale and bias of the BN layer to disentangle the normalization of the statistics of this layer.

\begin{figure*}[htbp]
\centering
\includegraphics[width=0.9\textwidth]{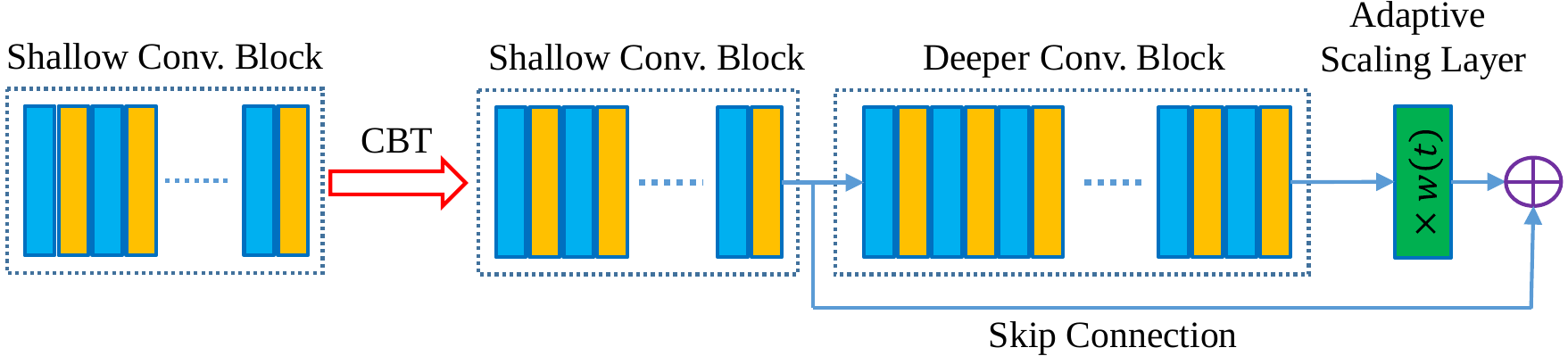}
\caption{Convolution Block Transformation (CBT) transfers weights of shallow network to the deeper one.}
\label{fig:cbt}
\end{figure*}

In Figure~\ref{fig:cbt}, we show the flowchart of the proposed CBT technique. With the help of CBT, to train a deeper model, we initialize a newly added convolution block (instead of a single layer) with Gaussian noise to break symmetry and add identity shortcut to preserve the potential ability of shallow model. As one can see in Figure~\ref{fig:cbt}, the weights of the shallow network are transferred to a consistent block of the deeper network. Some new convolution layers are added to the top of the shallow network. The output values of the newly added convolution block are scaled by an adaptive scaling layer. The adaptive layer is defined by the function $w(t)=1-e^{-t}$, where $t$ denotes the number of total iterations in one epoch divided by current iteration number. The adaptive scaled output is added with that of the shallow layer. Finally, the added results are fed into a global average pooling (GAP) layer (for more details about GAP, please refer to the Section~\ref{sec:training}). Here, we call the up-described operator as the convolution block transformation (CBT). Along with the training, the function $w(t)$ for the adaptive layer will gradually approach to $1$ and the newly added convolution block will becomes a part of the original shallow net.

\subsection{Feature Matching}
Generative Adversarial Networks (GANs) are difficult to train since the generator is easy to collapse~\cite{gan_tutorial} (we call it the ``model collapse'' phenomenon). In~\cite{improvegan},  in order to avoid mode collapse, Salimans et al. proposed the feature matching technique to improve the stability in training  the GANs by employing a new objective for the generator. Instead of maximizing the output of the discriminator as the regular GAN training does, the feature matching requires the generator to create latent data that matches the statistics of the realistic data in the feature level of the discriminator network. Consequently, the generator updates its parameters by matching the expectation of the features on the next of the final layer of the discriminator network, which is the output of Global Average Pooling (GAP) layer in our case. This is a natural choice of statistics for the generator to match. Let $f(x)$ denote activations after GAP layer of the discriminator, the feature matching objective for the generator is proposed by~\cite{improvegan} and defined by an $\ell_1$ distance as $||E_{x\in p_{data}}f(x)-E_{z\in p_z(z)} f(G(z))||_{1}$. In practice, we found that the above mentioned $\ell_1$ distance produces similar results with $\ell_2$ distance. The authors in~\cite{mmd_gan} proved that the maximum mean discrepancy (MMD) using Gaussian kernels could match all moment's mean, including the $\ell_1$ and $\ell_2$ distances between the generated features and unlabeled images. Therefore, in this work we employ the MMD metric as the feature matching objective to measure the distance between the features of generated images and unlabeled images. Then, we require generator to match the all levels statistics features of realistic data:
\begin{align}
\begin{split}
\textrm{MMD}(\mathcal{F},p_{data},p_z)&=\sup_{f \in \mathcal{F}}(\mathbb{E}_{x\sim p_{data}}[f(x)\\
&-\mathbb{E}_{z\sim p_z}[f(G(z))]),
\label{eq:MMD}
\end{split}
\end{align}
where $\mathcal{F}$ is a set of functions. When $\mathcal{F}$ is in a reproducing kernel Hilbert space (RKHS), the function approaching the supremum can be derived analytically and is called the witness function
\begin{align}
f(x) = \mathbb{E}_{x\sim p_{data}}[\mathcal{K}(x,G(z))] 
- \mathbb{E}_{z\sim p_z}[\mathcal{K}(x,G(z))],
\label{eq:witness}
\end{align}
where $\mathcal{K}$ is the kernel of the RKHS. Here, we assume $\mathcal{K}$ is measurable and bounded. Then we substitute \eqref{eq:witness} into \eqref{eq:MMD} and yield:
\begin{align}
\begin{split}
\textrm{MMD}^2(\mathcal{F},p_{data},p_z) &=
  \mathbb{E}_{x,x'\sim p_{data}}[\mathcal{K}(x,x')] \\
  &- 2\mathbb{E}_{x\sim p_{data},z\sim p_z}[\mathcal{K}(x,G(z))] \\
  &+ \mathbb{E}_{z,z\sim p_z}[\mathcal{K}(G(z),G(z'))].
\end{split}
\end{align}
This expression only involves expectations of the kernel $\mathcal{K}$, which can be approximated by:

\begin{align}
\begin{split}
  \textrm{MMD}_{sample}^2(\mathcal{F},p_{data},p_z) &= 
   \frac{1}{m^2}\sum_{i,j=1}^{m}\mathcal{K}(x_i,x_j)\\ 
- \frac{2}{mn}\sum_{i,j=1}^{m,n}\mathcal{K}(x_i,G(z_j))
  &+ \frac{1}{n^2}\sum_{i,j=1}^{n}\mathcal{K}(G(z_i),G(z_j))
\label{eq:MMD_b}
\end{split}
\end{align}
The MMD metric also depends on the choice of the kernel. We choose the inner product kernel for simplicity.
\begin{algorithm}
  \caption{Training of SGGAN.}
  \label{alg:selfgrow}
  \begin{algorithmic}[1]
  \For {$e=1,\ldots,epoches$}  
  \For {$t=1,\ldots,batches$}
  \State Generate images by using generator $G$.
  \State Feed generated, unlabeled, and labeled images into discriminator $D$ to obtain $Loss_{d}$.
	\State Compute $\frac{\partial Loss_{d}}{\partial W_{d}}$ and update $W_{d}$ with $W_{g}$ fixed.
    \State Feed unlabeled and generated images into $D$ to compute $Loss_{g}$.
    \State Compute $\frac{\partial Loss_{g}}{\partial W_{G}}$ through $D$ and update $W_{g}$ with $W_{d}$ fixed.
\\
\quad\ \textbf{end for}
   \EndFor
   \State Inference unlabeled images and create latent-labeled dataset by using discriminator $D$,
\\
\textbf{end for}
  \EndFor
  \State Initialize a deeper model by using CBT preservation technique.
  \end{algorithmic}
\end{algorithm}

\subsection{Learning Objective}
A key challenging in semi-supervised GANs is how to construct the loss function. For the losses, we find that the cross-entropy with Adam is a good choice for the optimizer. In~\cite{improvegan}, Salimans et al. introduce an effective strategy to construct the discriminator loss function $Loss_d$. They regard the labeled and unlabeled data as one of $k$ classes and then classify the latently generated data into the $(k+1)$-th class. In this way, $Loss_d$ can be defined as follows:
\begin{equation}
\begin{aligned}
Loss_d=&-\log(\frac{1}{\sum_{i=1}^m e^{g_i}+1})-\sum_{i=1}^m label_i\times\log(x_i)\\
&-\log(\frac{\sum_{i=1}^m e^{u_i}}{\sum_{i=1}^m e^{u_i}+1})
\label{eq:loss_d}
\end{aligned}
\end{equation}
where $-\sum_{i=1}^m label_i\times\log(x_i)$, $-\log(\frac{1}{\sum_{i=1}^m e^{g_i}+1})$, and $-\log(\frac{\sum_{i=1}^m e^{u_i}}{\sum_{i=1}^m e^{u_i}+1})$ are the losses related to generated, labeled, and unlabeled images, respectively. Here, $m$ is the batch size, and $x_i$, $g_i$, $u_i$ represent the output (before softmax activation) of the labeled, generated, and unlabeled images, respectively. During the training of the generator, a simple feature matching method is introduced to measure the dissimilarity between two distributions of realistic and latently generated data as described in~\cite{improvegan}. Motivated by the effectiveness of the maximum mean discrepancy (MMD)~\cite{mmd_origin,mmd_gan}, in the proposed SGGAN we utilize MMD metric instead of the $\ell_1$ to measure the dissimilarity between latently generated data and the realistic data.

\subsection{Training}
\label{sec:training}
A complete cycle of training the proposed SGGAN contains three iterative steps: 1) train the generator $G$ and discriminator $D$ on the labeled and unlabeled pool. Here, we employ the MMD metric for the updating of the weights of generator $G$; 2) apply the discriminator $D$ to predict the unlabeled pool, and then assign the most confident samples of all the $k$ classes to the labeled pool; 3) self-grow the discriminator $D$ and generator $G$ to be deeper and more powerful. The overall procedures of training the proposed SGGAN is summarized in Algorithm~\ref{alg:selfgrow}.

\subsubsection{Pre-Training}
The purpose of pre-training is to train the initial baby GAN cell. Inspired by the feature matching techniques introduced in the Improved GAN~\cite{improvegan}, the process of pre-training could solve the problems in training the discriminator. After this stage, we have a baby discriminator which achieves an accuracy of over $80\%$ on the testing set of the CelebA dataset~\cite{celeba}. To this end, we can make use of the trained baby discriminator to infer the latent labels from the unlabeled images.
We use Adam with initializing learning rate of $0.01$ to train both the generator $G$ and the discriminator $D$. The weights of baby generator and baby discriminator are initialized by using Xavier’s method~\cite{szegedy2015going}. In all experiments, the pixel values of the images are normalized to $[-1, 1]$.

\subsubsection{Label Inference}
As we mentioned in Section~\ref{sec:1}, inferring the latent labels of the unlabeled images is the most significant step in training semi-supervised learning models. One typical way to obtain the latent labels of unlabeled data is to hypothesize that the labels predicted by the initial classifier is credible. Under this circumstance, the label inference problem is tackled. However, there are two issues in this approach. The first one is that the initial classifier can be inaccurate towards unlabeled data, and leading wrong absorption of inaccurate data and thus assimilating noise into the training data. The second one is that as the initial classifier does well on the same class of data, adding this type of unlabeled data as latently labeled samples may make the classifier only memorize this specific type of data and cannot be generalized to other data types. How to solve these two issues is crucial to the success of a semi-supervised self-growing network. 

For the first issue, we can largely weaken its impact by only selecting the images in the unlabeled pool which have larger recognition probability than a pre-set threshold value $\alpha$, which can be determined by performing recognition experiments on the validation set of benchmark datasets (please refer to the experimental section for more details) through grid search strategy. In this work, we set $\alpha=0.98$. For the second issue, we initialize the junior network from the trained baby counterpart by introducing the proposed CBT preservation technique, and generalize the representational power of the classifier, accordingly. Comparing to the than the shallower network, a deeper network can potentially learn additional useful information from the latent labeled data. The improving performance of the Alexnet~\cite{krizhevsky2012imagenet} to the VGG~\cite{Simonyan14c}, and finally to the ResNet~\cite{he2016deep}, all demonstrates the great successes in the ILSVRC~\cite{ILSVRC15} challenge on the Imagenet Dataset~\cite{imagenet_cvpr09}. For example, VGG~\cite{Simonyan14c} uses $3\times3$ convolution to achieve deeper architecture and ResNet~\cite{he2016deep} treats convolution added with shortcut as a basic unit and repeats that unit until the depth limit of the network is reached. Going deeper can really improve the capacity of network considerably. As the model grows up stronger (deeper), the network can learn useful information not only on labeled images, but also on the latently labeled images.

\section{Experiments}
\label{sec:results}
In this section, we first evaluate the proposed semi-supervised self-growing GAN (SGGAN) approach and justify the effectiveness of each component in the SGGAN approach. Then we compare SGGAN with other state-of-the-art semi-supervised GAN based approaches on image recognition problem on two widely employed datasets. To demonstrate the broad applicability of the proposed SGGAN approach, we also compare it with the leading supervised deep learning approaches on two commonly used datasets for face attribute recognition.
\noindent 
\subsection{Dataset Description}
\begin{figure}[htbp]
\centering
\includegraphics[width=0.45\textwidth]{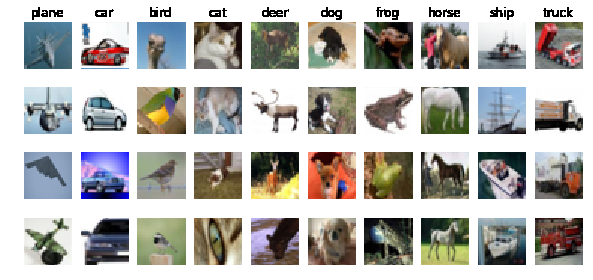}
\caption{Samples from the CIFAR-10 dataset~\cite{cifar}.}
\label{fig:cifar_data}
\end{figure}
\noindent \textbf{Image Recognition Datasets}. In this section, we compare the proposed SGGAN approach with state-of-the-art semi-supervised GAN based methods by using the widely used CIFAR-10 dataset~\cite{cifar} and the Street View House Numbers (SVHN) dataset~\cite{svhn}.

The CIFAR10 dataset~\cite{cifar} is introduced by A. Krizhevsky and G. Hinton in 2009, and has been a benchmark dataset for image classification problem ever since. This dataset contains $60,000$ color images of size $32\times32$ in 10 classes (i.e., airplane, automobile, bird, cat, deer, dog, frog, horse, ship, and truck). Some samples of this dataset are shown in Figure~\ref{fig:cifar_data}. Each class includes $6,000$ images, of which $5,000$ images are used for training and $1,000$ images are used for testing. It is a widely used dataset for evaluating both supervised and semi-supervised learning methods on the image recognition problem. We follow the same experimental setting as the previous work such as~\cite{improvegan}, in which only $100$, $200$, $400$ and $800$ samples along with their labels for each class are randomly selected as the training data for semi-supervised learning.

\begin{figure}[htbp]
\centering
\includegraphics[width=0.45\textwidth]{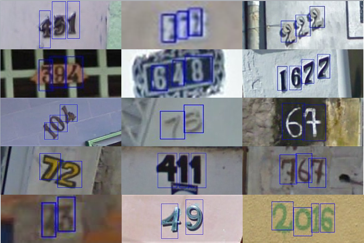}
\caption{Samples from the SVHN dataset~\cite{svhn}.}
\label{fig:svhn_data}
\end{figure}

The SVHN dataset~\cite{svhn} is a real-world image dataset for digit recognition problem. It is similar in flavor to the MNIST dataset~\cite{mnist}, but serves with a harder and real-world problem in the wild. This dataset contains over $600,000$ color digit images coming from the house numbers in Google Street View images. Some samples are listed in Figure~\ref{fig:svhn_data}. Among these images, there are $73,257$ images in the training set, $26,032$ images in the testing set. Following the experimental settings as described in~\cite{improvegan}, in which only $50$, $100$ and $200$ samples along with their labels for each class are selected as the training data for semi-supervised learning.

\noindent \textbf{Facial Attribute Recognition Datasets}. We also compare the proposed SGGAN approach with the leading supervised deep learning methods on facial attribute recognition problem with the CelebFaces Attributes Dataset (CelebA) dataset~\cite{celeba} and the Labeled Faces in the Wild-a (LFW-a) dataset~\cite{LFWTech}.

The CelebA dataset~\cite{celeba} is a large-scale face attributes dataset, which contains $202,599$ face images of $10,177$ identities in the wild, each of which includes $5$ landmark locations and $40$ binary attributes annotations. Among the $202,599$ face images in total, $19,962$ images are used as the testing set and the others are used as the training and validation set, respectively. In this article, we randomly select a small set of images as the training set and the others as the testing set.
\begin{figure}[htbp]
\centering
\includegraphics[width=0.46\textwidth, height=0.25\textwidth]{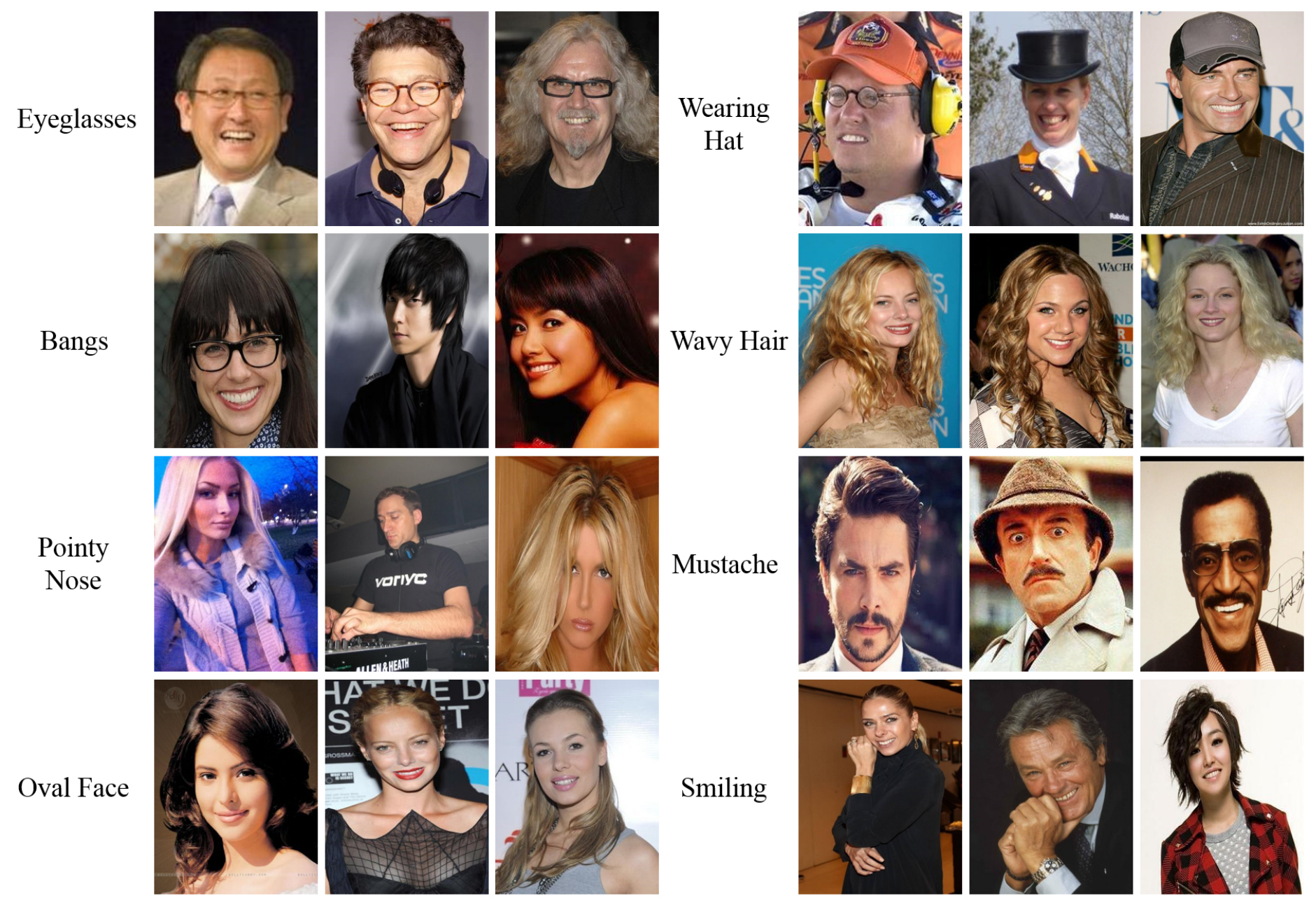}
\caption{Samples from CelebA dataset~\cite{celeba}.}
\label{fig:celeba_data}
\end{figure}
The LFWA dataset~\cite{LFWTech} has $13,233$ images of $5,749$ identities. Following the experimental settings as the previous work~\cite{celeba}, we employ $6,263$ images of $2,749$ peoples as the training set and the other $6,880$ images of $3,000$ peoples as the testing set. When we train the SGGAN model, the labeled images are randomly selected from the training set, and the final results on testing error are averaged by $10$ independent runnings. For the CelebA dataset~\cite{celeba}, the prediction threshold $\alpha$ is choose on the validation set.

In all these datasets (except the LFWA dataset~\cite{LFWTech}), we train the model on the training set and select the model trained with the lowest recognition error on the validation set, and report the testing error with the selected training model accordingly. For the LFWA dataset~\cite{LFWTech}, we follow the experimental settings as described in~\cite{celeba}.
\begin{figure}[htbp]
\centering
\includegraphics[width=0.4\textwidth]{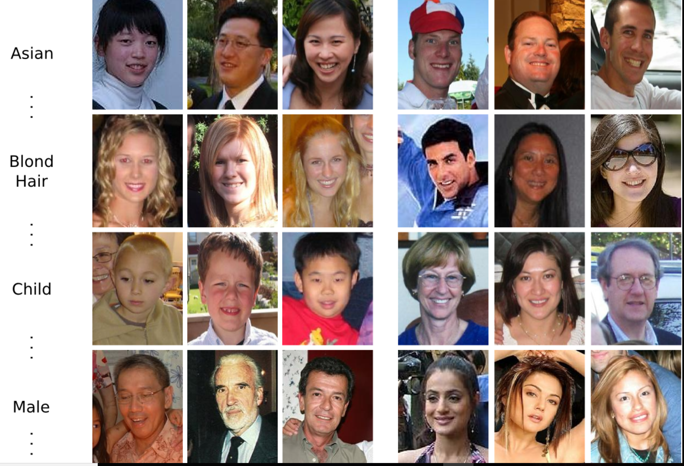}
\caption{Samples from LFWA dataset~\cite{LFWTech}.}
\label{fig:lfwa}
\end{figure}
\subsection{Ablation Study}
In this section, we justify the influence of different components in our proposed SGGAN approach on the performance of recognition errors. The aspects we investigate here include the network self-growing route, the objective function of the feature matching, the generated samples, and the comparison with transfer learning approaches, etc. All these study is evaluated on the training set of the CelebA dataset~\cite{celeba}.

\noindent\textbf{Network Self-Growing Route}. In our SGGAN model, the model is designed to \enquote{grow up} from a small one to a big one. However, how to decide the route for our SGGAN model to achieve better performance (i.e., lower recognition error) is still a big problem. The routes for the model to \enquote{grow up} can be very different. For example, the model can be directly grow from a baby model to a junior model, or from a baby model to a senior model, or from a baby model to a junior model and finally to a senior model, etc. To th
is end, we design a series of experiment to validate the most suitable \enquote{grow up} route for the proposed SGGAN model. 

We compare the proposed SGGAN model with different routes of \enquote{grow up} on the CelebA dataset~\cite{celeba}. The comparison is performed by using the \enquote{gender} attribute of $800$ labeled images. The experimental routes are summarized in the first three rows of the Table~\ref{tab:grow}, while symbol \enquote{$\surd$} indicates that the corresponding baby/junior/senior model is employed as a part of the whole training model and \enquote{$-$} indicates that the we skip the corresponding model. The order in models with three models is to train the whole model from baby one, junior one, to the senior one. From the last row of the Table~\ref{tab:grow}, one can see that the SGGAN model with the route of \enquote{grow up} along all the three models can achieve higher accuracy than with the other routes. Besides, the SGGAN model \enquote{grow up} with two models can achieve better performance than the SGGAN model with only one baby/junior/senior model. Similar findings can be found in the experiments on other attributes of the CelebA dataset~\cite{celeba} as well as on other datasets such as LFWA~\cite{LFWTech}. These results demonstrate that the network self-growing strategy can effectively improve the image recognition accuracy over the one with fixed single model. Specifically, using all these three models can significantly improve the recognition accuracy of the SGGAN model with only single individual baby/junior/senior network. % Note that the accuracy of the SGGAN model with single senior model achieves inferior performance to the junior model.

\begin{table}[htp!]
\caption{The accuracy (\%) of the proposed SGGAN network with different self-growing routes by using the \enquote{gender} attribute in the CelebA dataset~\cite{celeba}.}
  \label{results_different_routes}
  \centering
  \begin{tabular}{cccccccc}
    \toprule
    Baby    & $\surd$  & $-$  & $-$  & $\surd$  &  $\surd$ & $-$ & $\surd$ \\
    Junior  & $-$  &  $\surd$ & $-$  & $\surd$  &  $-$ & $\surd$  &  $\surd$    \\
    Senior & $-$  & $-$  & $\surd$  & $-$  & $\surd$  & $\surd$  &  $\surd$    \\
    \midrule
    Accuracy  & 80.2  & 85.1  & 84.6  & 86.7  & 88.5 & 89.1   & \textbf{89.6}     \\
    \bottomrule
  \end{tabular}
  \label{tab:grow}
  \end{table}

\noindent\textbf{Objective of Feature Matching}. The work of Wasserstein GAN~\cite{wgan} discusses different distances between distributions adopted by existing generative adversarial algorithms, and show many of them are discontinuous, such as Jensen-Shannon divergence~\cite{goodfellow2014generative} and Total Variation~\cite{total_variation},
except for Wasserstein distance. The discontinuity makes the gradient descent infeasible for training. Consequently,
~\cite{nips_mmd_gan} show Wasserstein
GAN~\cite{wgan} is a special case of the MMD, and hence MMD also has the advantages of being continuous and differentiable. We adopt the powerful MMD metric to our work to stabilize the training of generator.

\begin{figure}[htp!]
    \centering \includegraphics[width=0.4\textwidth]{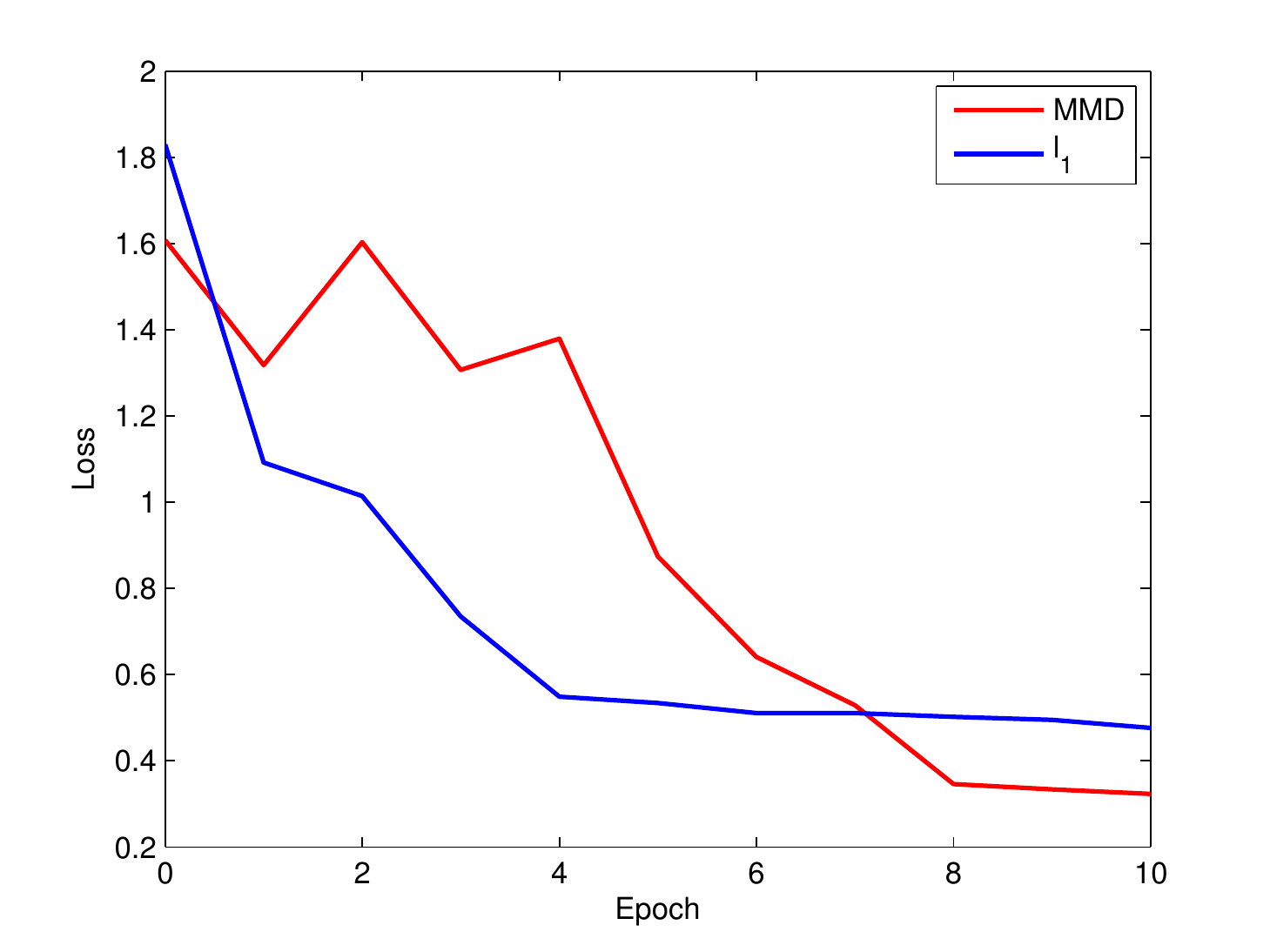}
    \caption{The loss function of the SGGAN model trained with MMD v.s. with $l_1$ distance as the objective of feature matching.}
    \label{fig:MMD}
\end{figure}

We compare the different objective of the feature matching step, i.e., the Maximum Mean Discrepancy (MMD) and the $\ell_{1}$ distance as we have mentioned in Section~\ref{sec:related_work}. Since the model collapse is a fundamental problem in the training of GAN, we use MMD to stabilize the GAN. The results are shown in Figure~\ref{fig:MMD}, from which one can see that the generator trained with the MMD objective can achieve lower training loss than that trained with $\ell_1$ distance after several epochs. This demonstrates that MMD is more suitable than the $\ell_1$ distance to be the loss objective function during the training of the generator in GAN.

\begin{figure}[htp!]
    \centering \includegraphics[width=0.4\textwidth]{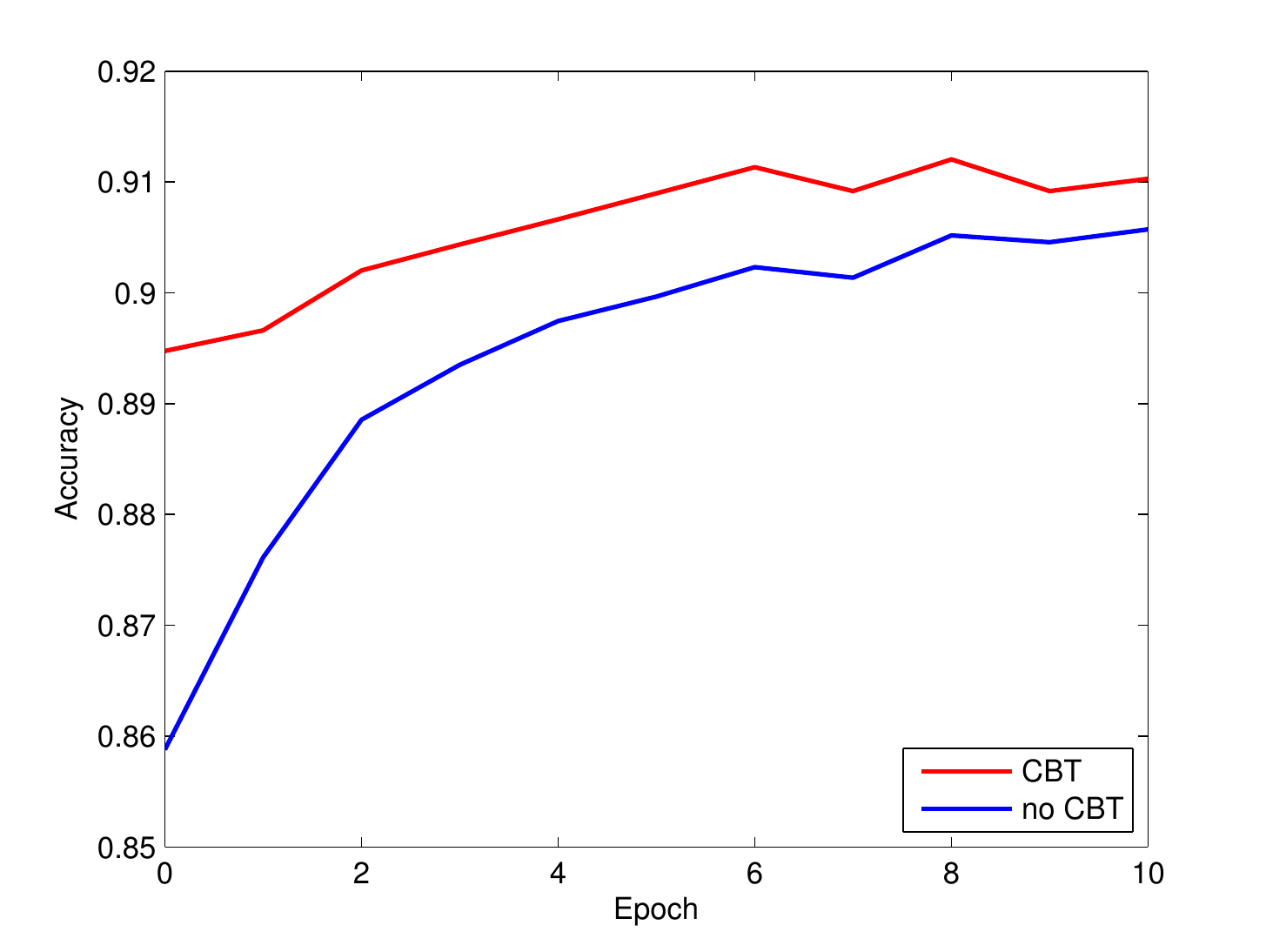}
    \caption{The loss function of the SGGAN model trained with the CBT v.s. without CBT during training.}
\label{fig:CBT}
\end{figure}

\noindent\textbf{Convolution Block Transformation (CBT)}. Figure~\ref{fig:CBT} shows that the recognition accuracy (\%) of the SGGAN model trained with the CBT technique are consistently higher than the model trained without CBT in different epochs. This demonstrate that CBT is more effective than its counterpart that simply copies the weights in the shallow model and initializes the newly added convolution-block layers randomly. This is due to the reason that the simple \enquote{training without CBT} strategy will wreck the weights in the shallow layers of the network. And evidence our proposed CBT technique will make the transfer of weights smoothly and hence preserve the function of shallower model at the beginning of training.

\noindent\textbf{Comparisons with Fine-tuned VGG and ResNet  Networks}. In order to show the advantages of our algorithm on label effectiveness, we compare our SGGAN model with the state-of-the-art networks such as the VGG-16 network~\cite{Simonyan14c} and the ResNet network~\cite{he2016deep} in the deep learning field. For the two networks, we load the model provided by corresponding authors pre-trained on the ImageNet dataset~\cite{imagenet_cvpr09} (which contains 1000 classes with 1.2 million images), and then carefully fine-tune these networks on the training set of the CelebA dataset~\cite{celeba}. The fine-tuning procedure can usually help the original networks yield better performance than training those networks on small dataset directly. The proposed SGGAN model, the pre-trained VGG-16 and Resnet-50 networks are all fine-tuned with different numbers (i.e., 800, 1600, 3200, 4800, 6400, 7200) of labeled images in the CelebA dataset~\cite{celeba} with "gender" attribute in the comparison experiments. We fine-tune the VGG-16 and ResNet-50 networks in a standard manner as described in corresponding paper. When the training set is of small scale, it is hard to train a very deep network from scratch. And the most frequently employed technique in literature is to fine-tune the off-the-shell networks, such as the famous VGG network~\cite{Simonyan14c}. We compare the proposed SGGAN approach with the fine-tuned VGG-16 and ResNet-50 networks with different numbers of labeled training images. 

The results on accuracy (\%) are listed in Table~\ref{tab:vgg}, from which one can see that when the numbers of labeled training images are $800$, $1,600$, $3,200$, and $4,800$, the proposed SGGAN approach can achieve higher recognition accuracies than the fine-tuned VGG-16 and ResNet-50 networks on the CelebA dataset with the \enquote{gender} attribute. Similar results can be found when we perform experiments on the other attributes of the CelebA dataset or the other datasets. When the numbers of the training samples increase to $6,400$ and $7,200$, the proposed SGGAN approach obtains slightly inferior (but still comparable) performance to the VGG-16 and ResNet-50 networks. All these results demonstrate the competing ability of the proposed SGGAN approach as a whole system over the leading VGG and ResNet networks on image recognition tasks such as face attribute recognition.
\begin{table}[htb!]
\caption{Comparison with the VGG-16 and ResNet-50 networks fine-tuned with different numbers of labeled images from the CelebA dataset~\cite{celeba} with the \enquote{gender} attribute.}
\centering
%\tiny
\label{tab:vgg}
\begin{tabular}{cccccccc}
\toprule
\multicolumn{1}{c}{\# of Labeled Image} & 800 & 1600 & 3200 & 4800 & 6400 & 7200  \\ 
\midrule
VGG16~\cite{Simonyan14c} & 89.3 & 92.4 & 94.8 & 95.9 & 97.6 & 98.1       \\ 
resnet50~\cite{he2016deep} & 88.6 & 91.9 & 94.6 & 96.2 & \textbf{97.8} & \textbf{98.3}       \\ 
SGGAN & \textbf{89.6} & \textbf{94.3} & \textbf{95.5} & \textbf{96.4} & 96.8 & 97.1     \\
\bottomrule 
\end{tabular}
\end{table}

\subsection{Comparison with state-of-the-art semi-supervised learning approaches on image recognition}\label{sec:cifar}
\subsubsection{Problem Description}
Image recognition problem is the task of assigning one label to an input image from a fixed set of categories. It is a fundamental problem in computer vision community. Image recognition has a large variety of practical applications, and is related to many other computer vision tasks such as object detection and segmentation.

\subsubsection{Comparison Methods}
We compare the proposed SGGAN approach with other competing semi-supervised learning approaches such as the Ladder Network~\cite{ladder}, which proposed to train the ladder network simultaneously minimize the sum of
supervised and unsupervised cost functions by back-propagation, avoiding the
need for layer-wise pre-training. And some leading GANs based approaches such as CatGAN~\cite{catgan}, which is based on an objective function
that trades-off mutual information between unlabeled examples and their predicted
categorical class distribution, against robustness of the classifier to an adversarial generative model. And the Improved GAN~\cite{improvegan}, which propose a technique called feature matching to address the instability of GANs by specifying a new objective for the generator to prevents it from overtraining on the current discriminator. Instead of directly maximizing the output of the discriminator, the new objective $\ell_1$ requires the generator to generate data that matches
the statistics of the real data. 
We compare these competing methods on the CIFAR10 dataset and the SVHN dataset~\cite{svhn}.

\subsubsection{Results and Discussions}
The experimental results are shown in Table~\ref{tab:cifar} and Table~\ref{tab:svhn}. It can be observed from Table~\ref{tab:cifar} that we achieve competitive results with the state-of-the-art on the two datasets. As the CIFAR10 dataset~\cite{cifar}, the SVHN dataset~\cite{svhn} is used for validating semi-supervised learning methods. Table~\ref{tab:svhn} shows the testing error for SVHN experiment. One can see that the more labeled samples we use, the better the recognition performance the proposed SGGAN model will be.

\begin{table}[htp!]
\caption{Comparison test error with other semi-supervised learning methods on CIFAR10 dataset. The results are averaged by 10 runs. N/A is not available, which is not report in their papers.}
\centering
\label{tab:cifar}
\begin{tabular}{cccccccc}
\toprule
\multicolumn{1}{c}{} & 1000 & 2000 & 4000 & 8000\\ 
\midrule
Ladder network~\cite{ladder} & N/A & N/A &20.4 &N/A\\
CatGAN~\cite{catgan} & N/A & N/A & 19.58 &N/A\\
Improved GANs~\cite{improvegan} & 21.83 & 19.61 & 18.63 & 17.72\\
\hline
SGGAN & \textbf{20.04} & \textbf{18.43} & \textbf{15.65} & \textbf{16.51}\\
\bottomrule 
\end{tabular}
\end{table}

\begin{table}[htp!]
\caption{Comparison test error with other semi-supervised learning methods on SVHN dataset. All experiments are averaged by $10$ runs.}
\centering
\label{tab:svhn}
\begin{tabular}{cccccccc}
\toprule
\multicolumn{1}{c}{} & 500 & 1000 & 2000   \\ 
\midrule
DGN~\cite{DGN} & 36.02 & N/A & N/A &\\
Virtual Adversarial~\cite{distributional} & 24.63& N/A & N/A &\\

Auxiliary Deep Generative Model~\cite{auxiliary} &22.86  & N/A & N/A &\\
Skip Deep Generative Model~\cite{auxiliary} & 16.61 & N/A & N/A &\\
Improved GANs~\cite{improvegan} & 18.44 & 8.11 & 6.16 \\ 
\hline
SGGAN & 17.31 & 6.53 & 5.13 \\
\bottomrule 
\end{tabular}
\end{table}

\subsection{Comparisons with supervised learning approaches on face attribute recognition} 
\begin{table*}[htp!]
\centering
\caption{Comparison with supervised learning methods.}
{
\begin{tabular}{l|l|l|l|l|l|l|l|l|l|l|l|l|l|l|l|l|l|l|l|l|l|l|}
\multicolumn{2}{r|}{}&\rotatebox{90}{5 o Clock Shadow}&\rotatebox{90}{Arched Eyebrows}&\rotatebox{90}{Attractive}&\rotatebox{90}{Bags Under Eyes}&\rotatebox{90}{Bald}&\rotatebox{90}{Bangs}&\rotatebox{90}{Big Lips}&\rotatebox{90}{Big Nose}&\rotatebox{90}{Black Hair}&\rotatebox{90}{Blond Hair}&\rotatebox{90}{Blurry}&\rotatebox{90}{Brown Hair}&\rotatebox{90}{Bushy Eyebrows}&\rotatebox{90}{Chubby}&\rotatebox{90}{Double Chin}&\rotatebox{90}{Eyeglasses}&\rotatebox{90}{Goatee}&\rotatebox{90}{Gray Hair}&\rotatebox{90}{Heavy Makeup}&\rotatebox{90}{H. Cheekbones}&\rotatebox{90}{Male}  \\ \hline
\multirow{5}{*}{CelebA}
&Face Tracer&85&76&78&76&89&88&64&74&70&80&81&60&80&86&88&98&93&90&85&84&91  \\ 
&PANDA-w&82&73&77&71&92&89&61&70&74&81&77&69&76&82&85&94&86&88&84&80&93  \\ 
&LNet+ANet(w/o)&88&74&77&73&95&92&66&75&84&91&80&78&85&86&88&96&92&93&85&84&94  \\ 
&LNet+ANet&\textbf{91}&\textbf{79}&\textbf{81}&\textbf{79}&98&\textbf{95}&\textbf{68}&78&\textbf{88}&\textbf{95}&\textbf{84}&80&\textbf{90}&\textbf{91}&92&\textbf{99}&\textbf{95}&\textbf{97}&\textbf{90}&\textbf{87}&\textbf{98}  \\  \hline
&Virtual GAN&84&73&75&71&92&90&62&74&80&90&77&76&82&85&89&92&88&91&85&80&91  \\ 
&Auxiliary GAN&85&73&75&74&93&91&63&75&83&91&80&77&83&84&90&93&91&90&86&83&92  \\ 
&Cat GAN&87&72&76&72&93&92&62&77&81&91&78&75&85&86&90&93&89&91&84&83&93  \\ 
&Skip GAN&
88&75&77&75&96&92&64&78&84&92&81&78&86&87&91&96&92&93&87&84&95  \\ 
&Improved GAN&87&76&78&76&95&91&65&79&85&91&82&79&87&88&91&95&90&92&88&86&93 \\ \hline

&SGGAN&90&77&79&77&\textbf{98}&94&66&\textbf{80}&86&94&83&\textbf{80}&88&89&\textbf{93}&98&94&95&89&86&97  \\ 
\hline
\multirow{5}{*}{LFWA}
&FaceTracer&70&67&71&65&77&72&68&73&76&88&73&62&67&67&70&90&69&78&88&77&84 \\
&PANDA-w&64&63&70&63&82&79&64&71&78&87&70&65&63&65&64&84&65&77&86&75&86 \\  
&LNets+ANet(w/o)&81&78&80&79&83&84&72&76&86&94&70&73&79&70&74&92&75&81&91&83&91 \\ 
&LNets+ANet&84&82&\textbf{83}&83&\textbf{88}&\textbf{88}&75&81&\textbf{90}&\textbf{97}&74&77&82&73&78&\textbf{95}&78&84&\textbf{95}&88&\textbf{94} \\  \hline
&Virtual GAN&80&79&77&81&82&81&71&77&84&91&73&74&78&70&74&89&76&80&89&84&88  \\ 
&Auxiliary GAN&81&80&78&82&83&82&72&78&85&92&74&75&79&71&75&90&77&81&90&85&89  \\ 
&Cat GAN&80&81&79&81&82&83&73&79&86&91&75&76&80&73&76&89&79&82&89&83&87 \\ 
&Skip GAN&82&83&81&83&86&83&73&81&88&93&75&78&82&72&78&91&90&82&93&80&90  \\ 
&Improved GAN&83&82&80&84&85&84&74&80&87&94&76&77&81&73&77&92&79&83&92&87&91  \\ \hline
&SGGAN&\textbf{85}&\textbf{84}&82&\textbf{86}&87&86&\textbf{76}&\textbf{82}&89&96&\textbf{77}&\textbf{79}&\textbf{83}&\textbf{75}&\textbf{79}&94&\textbf{81}&\textbf{85}&94&\textbf{89}&93  \\ \hline
\end{tabular}}
\label{tab:fully}
\end{table*}

\begin{table*}[htb!]
\centering
{
\begin{tabular}{l|l|l|l|l|l|l|l|l|l|l|l|l|l|l|l|l|l|l|l|l|l|l|}
\multicolumn{2}{r|}{}&\rotatebox{90}{Mouth S. O.}&\rotatebox{90}{Mustache}&\rotatebox{90}{Narrow Eyes}&\rotatebox{90}{No Beard}&\rotatebox{90}{Oval Face}&\rotatebox{90}{Pale Skin}&\rotatebox{90}{Pointy Nose}&\rotatebox{90}{Receding Hairline}&\rotatebox{90}{Rosy Cheeks}&\rotatebox{90}{Sideburns}&\rotatebox{90}{Smiling}&\rotatebox{90}{Straight Hair}&\rotatebox{90}{Wavy Hair}&\rotatebox{90}{Wearing Earrings}&\rotatebox{90}{Wear. Hat}&\rotatebox{90}{Wearing Lipstick}&\rotatebox{90}{Wearing Necklace}&\rotatebox{90}{Wearing Necktie}&\rotatebox{90}{Young}&\rotatebox{90}{\textbf{Average}} \\ \hline
\multirow{5}{*}{CelebA}
&Face Tracer&87&91&82&90&64&83&68&76&84&94&89&63&73&73&89&89&68&86&80&81\\ 
&PANDA-w&82&83&79&87&62&84&65&82&81&90&89&67&76&72&91&88&67&88&77&79  \\ 
&LNet+ANet(w/o)&86&91&77&92&63&87&70&85&87&91&88&69&75&78&96&90&68&86&83&83\\ 
&LNet+ANet&\textbf{92}&\textbf{95}&81&\textbf{95}&\textbf{66}&\textbf{91}&72&\textbf{89}&\textbf{90}&\textbf{96}&\textbf{92}&73&\textbf{80}&\textbf{82}&\textbf{99}&\textbf{93}&71&\textbf{93}&87&\textbf{87}  \\ \hline
&Virtual GAN&86&88&77&88&59&85&68&82&83&89&85&69&73&74&93&87&68&86&82&81  \\ 
&Auxiliary GAN&87&89&78&89&60&86&69&83&84&90&86&70&74&75&94&88&69&87&83&82  \\ 
&Cat GAN&88&88&77&88&61&87&68&82&85&91&87&71&75&74&93&87&68&76&84&83  \\ 
&Skip GAN&89&91&80&91&62&88&71&85&86&92&88&72&76&77&96&90&71&89&85&84  \\ 
&Improved GAN&90&90&81&90&63&89&72&84&87&91&87&73&77&76&95&91&72&88&84&85  \\ 
\hline

&SGGAN&91&93&\textbf{82}&93&64&90&\textbf{73}&87&88&94&90&\textbf{74}&78&79&98&92&\textbf{73}&91&\textbf{87}&86  \\ \hline
\hline
\multirow {5}{*}{LFWA}
&FaceTracer&77&83&73&69&66&70&74&63&70&71&78&67&62&88&75&87&81&71&80&74 \\
&PANDA-w&74&77&68&63&64&64&68&61&64&68&77&68&63&85&78&83&79&70&76&71 \\
&LNets+ANet(w/o)&78&87&77&75&71&81&76&81&72&72&88&71&73&90&84&92&83&76&82&79 \\
&LNets+ANet&82&\textbf{92}&81&79&74&\textbf{84}&80&85&78&77&\textbf{91}&76&76&\textbf{94}&88&95&88&79&\textbf{86}&84 \\  \hline
&Auxiliary GAN&78&86&78&77&71&78&76&82&75&76&85&75&74&88&84&91&85&76&78&80  \\ 
&Auxiliary GAN&79&87&79&78&72&79&77&83&76&77&86&76&75&89&85&92&86&77&79&81  \\ 
&Cat GAN&80&88&78&77&73&78&78&82&77&78&85&75&76&88&86&91&87&78&78&81  \\ 
&Skip GAN&81&89&81&80&74&81&79&85&78&79&88&78&77&91&87&94&88&79&81&83 \\ 

&Improved GAN&82&88&82&81&73&80&80&84&79&80&87&77&78&90&88&93&89&80&82&83  \\ \hline

&SGGAN&\textbf{83}&91&\textbf{83}&\textbf{82}&\textbf{76}&83&\textbf{81}&\textbf{87}&\textbf{80}&\textbf{81}&90&\textbf{80}&\textbf{79}&93&\textbf{89}&\textbf{96}&\textbf{90}&\textbf{81}&83&\textbf{85}  \\ \hline
\end{tabular}}
\end{table*}
\subsubsection{Problem Description}
Face attributes recognition is to get descriptive attributes on faces (gender, sex, the presence of sunglasses etc). Kumar et al.~\cite{kumar2009attribute} first introduced them as mid-level features for face verification~\cite{kumar2011describable} and since then have attracted much attention. The recognition of face attributes has an important role in computer vision applications due to their detailed description of human faces. The applications of it include suspect identification~\cite{klare2014suspect}, face verification~\cite{kumar2009attribute} and face retrieval~\cite{kumar2011describable}. Predicting face attributes in the wild in challenging due to complex face variations. In facial attribute recognition field, labeled data are either expensive or unavailable to obtain. Consequently, the large number of unlabeled face images available on the Internet have attracted increasing interests of researchers to tackle facial attribute recognition problem by semi-supervised learning (SSL)~\cite{Chapelle:2010:SL:1841234} methods.

\subsubsection{Comparisons methods}
The proposed method is compared with four competitive fully-supervised approaches including FaceTracer~\cite{kumar2008facetracer}, PANDA-w~\cite{zhang2014panda}, LNet+ANet(w/o) and LNet+ANet~\cite{zhang2014panda} on the two datasets mentioned above.
Compared with the fully-supervised learning methods, our self-growing approach only uses $7200$ labeled images. The LFWA dataset is a standard benchmark for face attribute classification. However, the number of training and validation data of LFWA data set is small, which made it not suitable to our algorithm. So we report two patterns of result on LFWA dataset. The first one uses all the training/validation data in the LFWA dataset and the other uses the data of CelebA as the unlabeled data pool. Our algorithm runs ten times, and we report the average result. 
%Every method uses all of training data.

\subsubsection{Results and Discussions}
The comparison results on CelebA and LFWA datasets are shown in Table~\ref{tab:fully}, from which one can see that the proposed SGGAN approach achieve comparable performance on the recognition accuracy when compared with the state-of-the-art supervised learning based deep learning methods. For example, the proposed SGGAN trained with the MMD objective and the CBT technique (i.e., SGGAN-MMD-CBT) achieves an accuracy of $86.22\%$, which is only slightly inferior to the LNet+ANet methods, but still superior to all the other methods. Note that the proposed SGGAN-MMD-CBT approach achieves such promising performance with only $4\%$ labeled training images.
\begin{figure*}[htp!]
\centering
\subfigure{
}
\subfigure{
\begin{minipage}[t]{0.25\textwidth}
\centering
\raisebox{-0.15cm}{\includegraphics[width=1\textwidth]{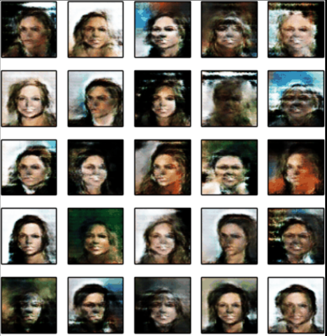}}
{\footnotesize Baby generator}
\end{minipage}
\hspace{6mm}
\begin{minipage}[t]{0.25\textwidth}
\centering
\raisebox{-0.15cm}{\includegraphics[width=1\textwidth]{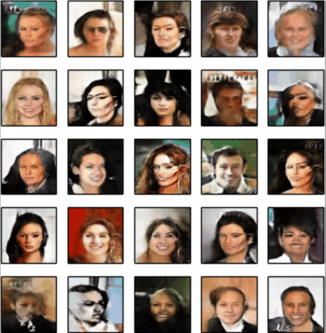}}
{\footnotesize Junior generator}
\end{minipage}
\hspace{6mm}
\begin{minipage}[t]{0.25\textwidth}
\centering
\raisebox{-0.15cm}{\includegraphics[width=1\textwidth]{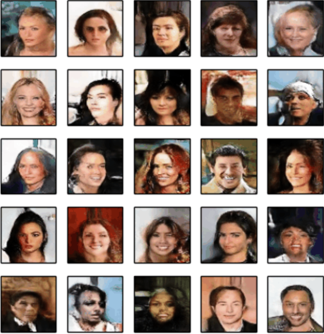}}
{\footnotesize Senior generator}
\end{minipage}}
\caption{The generated samples of the baby, junior, and senior generators of the proposed SGGAN approach.}
\label{fig:generated}
\end{figure*}

\subsection{Results on the LFWA dataset using external unlabeled data}
The LFWA dataset is a standard benchmark for face attribute recognition. However, the number of training images in LFWA dataset is small ($6,263$ images), which made it not well suitable for our algorithm. In order to achieve semi-supervised learning on LFWA dataset. We use all training images in CelebA dataset as the unlabeled pool for our algorithm to train a SGGAN on LFWA dataset. During training, the labels of CelebA dataset was not used. Table~\ref{tab:lfwa} show the results, in the first row of the table, the \enquote{LFWA (Outer data)} shows the result of SGGAN using CelebA as the unlabeled pool and \enquote{LFWA} is the result of SGGAN only use the images in LFWA. From the table one can see a large number of unlabeled images will improve around $6\%$ points for LFWA dataset. Which also demonstrate the effectiveness of our algorithm for the semi-supervised image recognition tasks.

\begin{table}[htp!]
  \caption{Recognition accuracy (\%) of SGGAN with different component settings on the LFWA dataset~\cite{LFWTech} by using unlabeled data.}
  \centering
  \begin{tabular}{lccc}
  \toprule
  Methods & LFWA (Outer data) & LFWA\\
\midrule
    Baseline: Feature Matching~\cite{improvegan} & 81.29  &  78.56  \\
    SGGAN-MMD &  83.41 &   78.53 \\
    SGGAN-CBT &  84.27 &   78.57   \\
    SGGAN-MMD-CBT &  85.32 &   78.81  \\
    \bottomrule
  \end{tabular}
\label{tab:lfwa}
\end{table}

\subsection{Generated Samples}
Feature matching is proved to help the GANs work much better if the goal is to obtain a strong classifier using the approach to semi-supervised learning~\cite{improvegan}. It works well for semi-supervised learning approaches. However, the samples generated by the generator during semi-supervised learning using feature matching do not look visually appealing. The reason appears to be that the human visual system is strongly attuned to image statistics that can help infer what class of object an image represents, while it is presumably less sensitive to local statistics that are less important for interpretation of the image. This is supported by the high correlation between the quality reported by human annotators and the Inception
score developed in the work~\cite{improvegan}. 

We show the generated samples in Figure~\ref{fig:generated}, from which one can see that the junior generator can produce samples with better image quality than those generated by the baby one, and the senior generator can further enhance the performance of the image quality of the produced samples than those generated by the junior one. This demonstrate that the proposed SGGAN network with \enquote{grow up} strategy can indeed make better generation during the training process, and hence implicitly help improve the performance of the GAN model on the recognition tasks.

\section{Conclusion}
\label{sec:conclusion}
In this paper, we propose a simple yet effective semi-supervised self-growing generative adversarial network (SGGAN) for image recognition. We propose a convolution-block-transformation (CBT) preservation technique to promote the network self-growing and obtain deeper network. Meanwhile, we leverage a maximum mean discrepancy (MMD) metric to stabilize and improve the training of SGGAN. The experiments on CIFAR10 and SVHN dataset demonstrate effectiveness our methods.  Extensive experiments on the CelebA and LFWA demonstrate the generalization of our method. With only around $4\%$ labeled training data, our SGGAN can achieve comparable performance with the fully-supervised convolutional neural network.

\balance
\bibliographystyle{plain}
\bibliography{main}
\end{document}